\documentclass[runningheads]{llncs}

\usepackage{eccv}

\usepackage{eccvabbrv}

\usepackage{graphicx}
\usepackage{booktabs}

\usepackage{url}            
\usepackage{booktabs}       
\usepackage{amsfonts}       
\usepackage{nicefrac}       
\usepackage{microtype}      
\usepackage{xcolor}         
\usepackage{graphicx}
\usepackage{amsmath}
\usepackage{amssymb}
\usepackage{amsfonts}
\usepackage{multirow}
\usepackage{color}
\usepackage{colortbl}
\usepackage{wrapfig}
\usepackage{balance}
\usepackage{paralist}
\usepackage{authblk}

\usepackage[accsupp]{axessibility}  

\usepackage{hyperref}

\usepackage{orcidlink}

\definecolor{mygray}{gray}{.96}
\definecolor{myblue}{RGB}{240,248,255}
\definecolor{myred}{RGB}{255,228,225}

\def\eg{\emph{e.g.}}

\begin{document}


\title{Current Challenges on Fine-grained Human Action Recognition}
\title{Comparisons of Vision Foundation Models on Fine-grained Human Action Recognition}
\title{Can Current Visual-language Models Fix Fine-grained Human Action Recognition?}
\title{A Study on Visual-language Models for Fine-grained Action Recognition}
\title{Are Visual-Language Models Effective in Action Recognition? A Comparative Study}
\titlerunning{Abbreviated paper title}

\author{Mahmoud Ali$^*$ \and
Di Yang$^*$  
\and
François Brémond}

\authorrunning{M. Ali, D. Yang and F. Brémond.}

\institute{Inria Center at Université Côte d'Azur, France \\
\email{\{mahmoud.ali, di.yang, francois.bremond\}@inria.fr}}

\maketitle

\def\thefootnote{*}\footnotetext{Equal contribution.}

\begin{abstract}
Current vision-language foundation models, such as CLIP, have recently shown significant improvement in performance across various downstream tasks. However, whether such foundation models significantly improve more complex fine-grained action recognition tasks is still an open question. To answer this question and better find out the future research direction on human behavior analysis in-the-wild, this paper provides a large-scale study and insight on current state-of-the-art vision foundation models by comparing their transfer ability onto zero-shot and frame-wise action recognition tasks. 
Extensive experiments are conducted on recent fine-grained, human-centric action recognition datasets (\eg, Toyota Smarthome, Penn Action, UAV-Human, TSU, Charades) including action classification and segmentation.

  \keywords{Video understanding \and video foundation model \and action recognition \and multi-modal learning }
\end{abstract}

\section{Introduction}

Recent vision-language foundation models~\cite{radford2021clip, luo2021clip4clip, xu2021vlm-videoclip, shukor2023epalm, shukor2023unified, yang2023vid2seq, Cadene_2019_CVPR, mezghani2023think-karteek-language, shukor2023unival, wang2022allinone, girdhar2023imagebind}, with a large-scale pre-training, have achieved promising results on many downstream vision tasks due to their impressive generalizability. Among these, the models with visual-language pre-training like CLIP~\cite{radford2021clip} and its successors for video tasks~\cite{Ma2022XCLIP, wang2023internvid, hanoonavificlip} have revolutionized a myriad of downstream tasks, demonstrating unprecedented versatility and performance. 

Despite these successes, the evaluations are primarily focused on general video understanding tasks such as video captioning, video-text retrieval, etc. The capability of these visual-language models to handle more complex and fine-grained action understanding tasks remains under-explored, such as zero-shot action classification and multi-label action segmentation. As these tasks are critical for many applications (\eg, healthcare monitoring and robotic learning), it is essential to understand the current challenges of visual-language models targeting fine-grained human action recognition. Hence, in this paper, we evaluate and compare current state-of-the-art (SoTA) visual-language models, with a particular focus on their performance in zero-shot classification and action segmentation tasks.

To further understand how to take good advantage of video-language model for zero-shot action recognition, we firstly compare different kinds of action descriptions obtained from raw action labels and LLMs (\eg, ChatGPT) for action classification to find out which kind of prompt is more appropriate for visual-language models. Secondly, for zero-shot action segmentation in untrimmed video, we apply current video question answering (VQA) models~\cite{Ren2023TimeChat, lin2023univtg} with post-processing for generating frame-level action predictions. We conduct a comparative study on such methods and classical action segmentation methods on more challenging multi-label dataset~\cite{Dai_2022_PAMI} to fully understand the advantages and limitations of each current approach.

In summary, the contributions of this paper are the following.
\begin{inparaenum}[(i)]
\item We perform a large-scale study on evaluating current vision-language foundation models focusing on transfer-learning onto in-the-wild action recognition tasks.  
\item We further provide insight and comparisons on different action description generation strategies for zero-shot action classification, and on different frame-wise action prediction strategies using video question answering (VQA) models for zero-shot action segmentation.
\item Extensive experiments are conducted using a good number of in-the-wild benchmarks.
\end{inparaenum}

\begin{figure}[t]
\begin{center}
\includegraphics[width=0.72\linewidth]{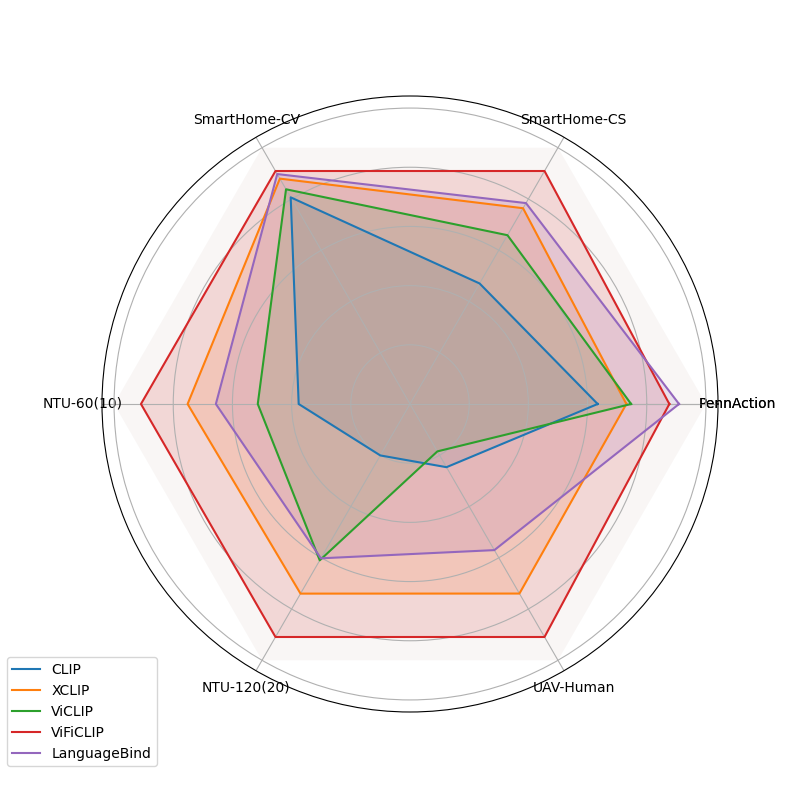}
\end{center}
   \vspace{-0.75cm}
   \caption{Statistics of the results on different datasets.}
\vspace{-0.35cm}
\label{fig:datasets}
\end{figure}
  
\section{SoTA Multi-modal Video Foundation Models}

\begin{table}[t]
\centering

\begin{center}
\scalebox{.75}{
\setlength{\tabcolsep}{0.65mm}{
\begin{tabular}{ l c c c c }
\hline
{\textbf{Methods}}& {\textbf{Training Data}}& {\textbf{Backbone}} & {\textbf{parameters (M)}}& \textbf{Training Strategy}\\

\hline
\hline
\text{CLIP~\cite{radford2021clip}} & CLIP-400-M/LAION-2B&  ResNet/ ViT& 151.3& image-text CL\\
\text{X-CLIP~\cite{Ma2022XCLIP}} & CLIP-400M/Kinetics-400 & Vit-B &  131.5 & video-text CL\\
\text{ViCLIP~\cite{wang2023internvid}} &InternVid-10M-FLT&Vit-L & 149.7& video-text CL\\
\text{ViFi-CLIP~\cite{hanoonavificlip}}&CLIP-400M/Kinetics-400 & ViT-B& 124.7 & video-text CL\\
\text{LanguageBind~\cite{zhu2023languagebind}}&VIDAL-10M & Vit-L& 528& image/audio/video-text CL\\

\hline

\end{tabular}}}
\end{center}
\vspace{-0.15cm}
\caption{A survey of SoTA architectures, CL: Contrastive Learning.}
\vspace{-0.65cm}
\label{tab_methods}
\end{table}

Recently, many methods have used language features~\cite{radford2021clip} for video understanding~\cite{luo2021clip4clip, xu2021vlm-videoclip, shukor2023epalm, shukor2023unified, mezghani2023think-karteek-language, wang2022allinone}, video captioning~\cite{yang2023vid2seq} and visual question answering~\cite{Cadene_2019_CVPR, shukor2023unival}. However, these methods are designed to handle short temporal videos, and the challenge of handling actions over a long range of time for solving the task of action detection still persists. These models, especially InternVideo~\cite{wang2022internvideo}, aim to understand and generate descriptions of video content, facilitating a multi-modal understanding of visual data. 

In this study, we select the most recent and widely used models~\cite{radford2021clip, Ma2022XCLIP, wang2023internvid, hanoonavificlip, Ren2023TimeChat} (see Tab.~\ref{tab_methods}) for comparisons and discussions.

\vspace{0.2cm}

\noindent\textbf{CLIP}~\cite{radford2021clip} is the first well known vision foundation model via visual-language pre-training. The key idea is to pre-train a transferable vision encoder using natural language supervision. The vision encoder is trained on a large number of image-text pairs by contrastive learning. CLIP successfully demonstrates that semantic information can significantly improve the representation ability of visual encoder for many downstream vision tasks, e.g., image classification, object detection. However, as the visual encoder of CLIP is trained based on images but not videos, CLIP is still limited on performance when transferred onto video-based tasks highly relying on temporal reasoning. In this study, we evaluate CLIP for video-based action recognition tasks as baseline model.

\vspace{0.2cm}

\noindent\textbf{X-CLIP}~\cite{Ma2022XCLIP} presents a novel multi-grained contrastive model for video-text retrieval. To effectively aggregate fine-grained and cross-grained similarity matrices to instance-level similarity, X-CLIP proposes the Attention Over Similarity Matrix (AOSM) module to make the model focus on the contrast between essential frames and words, thus lowering the impact of unnecessary frames and words on retrieval results. With multi-grained contrast and the proposed AOSM module, X-CLIP achieves outstanding performance on video-text retrieval tasks. In this work, we evaluate and compare X-CLIP focusing on fine-grained action recognition tasks with other SoTA CLIP-based approaches.

\vspace{0.2cm}

\noindent\textbf{ViCLIP}~\cite{wang2023internvid} is a general video foundation model.
It applies the Vision Transformer (ViT)~\cite{vit} with spatio-temporal attention as the video encoder and uses a Transformer-based text encoder following~\cite{radford2021clip}. It develops its capabilities through a mix of self-supervised methods, including masked modeling~\cite{tong2022videomae} and cross-modal contrastive learning~\cite{infonce} for in-depth feature representation, allowing for efficient learning of transferable video-language representation. As the video and text encoders are well pre-trained on a web-scale video-language dataset~\cite{wang2023internvid} including 7 million videos, corresponding to 234 million clips each with the generated captions, ViCLIP can be used for video and text feature extractions.

\vspace{0.2cm}

\noindent\textbf{ViFi-CLIP}~\cite{hanoonavificlip} explores the capability of a simple baseline called ViFi-CLIP (Video Fine-tuned CLIP) for adapting image pretrained CLIP to the video domain. ViFi-CLIP tackles the challenge of missing temporal relationships of image-based CLIP model, which can effectively improve the video-based downstream tasks. In this paper, we further evaluate this method for more fine-grained tasks.

\vspace{0.2cm}

\noindent\textbf{LanguageBind}~\cite{zhu2023languagebind} is a multimodal model that primarily uses language to connect different data types, like videos, infrared images, depth maps, and audio using contrastive learning. It is trained on a large-scale dataset (VIDAL-10M) which contains 10 million samples for all these data types and their corresponding text descriptions. To enhance the model's understanding and the language semantic information, the text descriptions are improved by incorporating metadata, spatial, and temporal information. Additionally, ChatGPT is used to refine the language and create a better semantic representation for each data type.

\vspace{0.2cm}

\noindent\textbf{TimeChat}~\cite{Ren2023TimeChat} is a time-sensitive multi-modal large language model specifically designed for long video understanding. 
Timechat is trained on an instruction-tuning dataset, encompassing 6 tasks and a total of 125K instances. This model shows promising zero-shot results on video understanding tasks including dense captioning, temporal grounding, and highlight detection. As this model can directly provide action segmentation prediction by asking related questions without additional training on downstream datasets, we compare this model to very challenging action segmentation tasks to understand its generalization ability. 

\vspace{0.2cm}

\noindent\textbf{UniVTG}~\cite{lin2023univtg} proposes to Unify the diverse Video Temporal Grounding (VTG) labels and tasks. Thanks to the unified framework, the temporal grounding pre-training is available from large-scale diverse labels and develops stronger grounding abilities \eg, zero-shot grounding. Similar as TimeChat, the zero-shot grounding can provide event boundaries related to action, hence, UniVTG can be used for zero-shot action detection segmentation tasks. In this study, we are the first to provide experimental results using UniVTG for more complex multi-label and frame-wise action segmentation tasks.

\vspace{0.2cm}

All mentioned approaches achieve SoTA performance on many tasks including video-text retrieval, temporal grounding, video captioning, etc. Most tasks are based on web videos and highly relys on video-text alignment quality, while are not focused on daily living action recognition scenarios. It is critical to understand the performance and current challenges of SoTA foundation models for action recognition tasks, so we provide an analysis on this topic to find out more future directions based on the analysis.

\section{Current Challenges on Action Recognition}
In this work, we provide an analysis of the performance of current vision foundation models with two challenging video-based tasks: zero-shot action classification and frame-wise temporal action segmentation. The evaluation and comparisons are performed on real-world datasets.

\subsection{Zero-shot Action Classification}
Zero-shot action classification is to pre-train an action classification model and then transfer this model onto an unseen dataset. Unlike traditional methods~\cite{Das_2019_ICCV, unik, das2021vpn+, yang2023ltn, yang2022via} that rely on extensive action labels, zero-shot approaches aim to generalize knowledge from known actions to unknown ones. Specifically, the semantic information, such as textual descriptions of the action labels, and the videos in the dataset are embedded using CLIP-based methods~\cite{cliporder, Ma2022XCLIP, wang2023internvid, hanoonavificlip}. Subsequently, given a video embedding, we search for its closest semantic information as the action prediction. We select such tasks as it highly relys on video-text alignment but has not been fully evaluated by current research.

In real-world video understanding applications, the ability to recognize actions without the need for specific training data is invaluable. However, visual features are often low-level, such as shapes, colors, and motions, while action descriptions are more abstract, this makes the model difficult to accurately match the two types of features. Additionally, current zero-shot learning models are still limited to dealing with variations in camera angles, lighting conditions, etc. Hence, this study aims to evaluate and compare the CLIP-based vision foundation models on such tasks focusing on real-world scenarios.

\vspace{0.2cm}

\subsection{Frame-wise Action Segmentation in Untrimmed Videos}
Temporal Action Segmentation focuses on per-frame activity classification in untrimmed videos~\cite{dai2022mstct, yang2023lac}. The main challenge is how to model long-term relationships among various activities at different time steps. 
Specifically,
action segmentation entails the automatic partitioning of untrimmed video sequences into distinct segments, each corresponding to a coherent action. Current methods~\cite{dai2021pdan, dai2022mstct} have two steps, they firstly extract visual features on top of the temporal segments of a long-term video using a strong video encoder. Secondly, they design temporal modeling to process the features. Hence, the performance of the temporal modeling highly relies on the video encoder from current video foundation models. In this study, we compare SoTA vision foundation models~\cite{radford2021clip, Ma2022XCLIP, wang2023internvid, hanoonavificlip} by evaluating their features on temporal action segmentation tasks.

\subsection{Evaluation Datasets}
\begin{table*}[t]
\centering

\begin{center}

\scalebox{0.8}{
\setlength{\tabcolsep}{1.mm}{
\begin{tabular}{  l c c c c c c c}
\hline

\textbf{Dataset}& \textbf{Real-world}&\textbf{2D }&\textbf{3D} &\textbf{\#Videos}&\textbf{\#Actions}&\textbf{Fine-grained} &\textbf{Type}\\ 
\hline
\hline

NTU-RGB+D 60~\cite{Shahroudy2016NTURA} &$\times$ &\checkmark &\checkmark &56,880 &60 & No & Daily living\\ 
NTU-RGB+D 120~\cite{NTU-120} &$\times$ &\checkmark &\checkmark &114,480 &120 & No & Daily living\\ 
Penn Action~\cite{penn}&\checkmark &\checkmark &$\times$ &2,326 &15 & No & Sport\\

UAV-Human~\cite{uav}
&\checkmark &\checkmark &$\times$ & 21,224& 155 & No  &  UAV\\
Toyota Smarthome~\cite{Das_2019_ICCV} 
&\checkmark &\checkmark &\checkmark &16,115 &31 & Yes  & Daily living
\\ 
Kinetics~\cite{Carreira_2017_CVPR} 
&\checkmark& $\times$ & $\times$& 400,000 & 400 & No  & General video
\\

\hline
PKU-MMD~\cite{liu2017pku} 
&$\times$ &\checkmark &\checkmark &1,076 &51 & No  & Daily living
\\
Charades~\cite{Sigurdsson2016HollywoodIH} 
&\checkmark &$\times$  &$\times$  &2,300 &151 & Yes  & Daily living
\\ 
TSU~\cite{Dai_2022_PAMI} 
&\checkmark &\checkmark &\checkmark &536 &51 & Yes&  Daily living
\\ 

\hline

\end{tabular}
}}

\end{center}
\vspace{-0.15cm}
\caption{A survey of recent datasets for in-the-wild human action classification (top), action segmentation (bottom).}
\vspace{-0.65cm}
\label{tab:dataset}
\end{table*}

Tab.~\ref{tab:dataset} summarizes the current challenging datasets targeting human behavior analysis. In this paper, we focus on two current challenging tasks, zero-shot classification and frame-wise segmentation tasks. Specifically, we perform the study on real-world scenarios~\cite{Das_2019_ICCV, uav, penn, Dai_2022_PAMI, Sigurdsson2016HollywoodIH} and laboratory scenarios~\cite{Shahroudy2016NTURA, NTU-120} for action understanding including both zero-shot classification and frame-wise segmentation tasks.

\noindent\textbf{Toyota Smarthome} (Smarthome)~\cite{Das_2019_ICCV} is a real-world human-centric daily living action classification dataset. The dataset is challenging as the inter-class variance is small and the activities are fine-grained. It contains 16,115 videos across 31 action classes, offering RGB and skeleton data. We utilize only RGB data, following cross-subject (CS) and cross-view2 (CV2) protocols and we report Top-1 accuracy in this work.

\noindent\textbf{UAV-Human}~\cite{uav} features 22,476 UAV-captured human-centric videos, we use the RGB data and follow Cross-subject evaluations (CS1).

\noindent\textbf{Penn Action}~\cite{penn} comprises 2,326 sequences of 15 simple sport actions, we use this dataset for action classification using standard train-test splits.

\noindent\textbf{NTU-RGB+D 60}~\cite{Shahroudy2016NTURA} includes 60 indoor daily living activities and consists of 56,880 RGB-D video sequences with 3D skeletons, captured by the Microsoft Kinect v2 sensor. We only use RGB videos in this work and we follow the cross-subject (CS) evaluation protocol.

\noindent\textbf{NTU-RGB+D 120}~\cite{NTU-120} extends the number of action classes and videos of NTU-RGB+D 60 to 120 classes 114,480 videos. We follow the cross-subject (CS) evaluation protocols. 

\vspace{0.2cm}
\noindent\textbf{Toyota Smarthome Untrimmed} (TSU)~\cite{Dai_2022_PAMI} extends the action classes and video counts of Smarthome, focusing on frame-wise segmentation tasks. The dataset is very challenging, as each action can be performed multiple times in a video and multiple actions can be performed at the same time. We use TSU for evaluating the generalizability of SoTA models and we report per-frame mAP following Cross-Subject (CS) and Cross-View (CV) evaluation protocols.

\noindent\textbf{Charades}~\cite{Sigurdsson2016HollywoodIH} focuses on fine-grained activities segmentation. It contains many object-oriented activities and variant light conditions. The current methods are still limited to dealing with this dataset, hence, we use this dataset for our study and we report per-frame mAP. 

\vspace{0.2cm}
The mentioned datasets are different from the datasets of web videos used for training video foundation models. Our selected evaluated datasets can further reflect the generalization ability of video foundation models on daily living scenarios.

\begin{figure}[t]
\begin{center}
\includegraphics[width=1\linewidth]{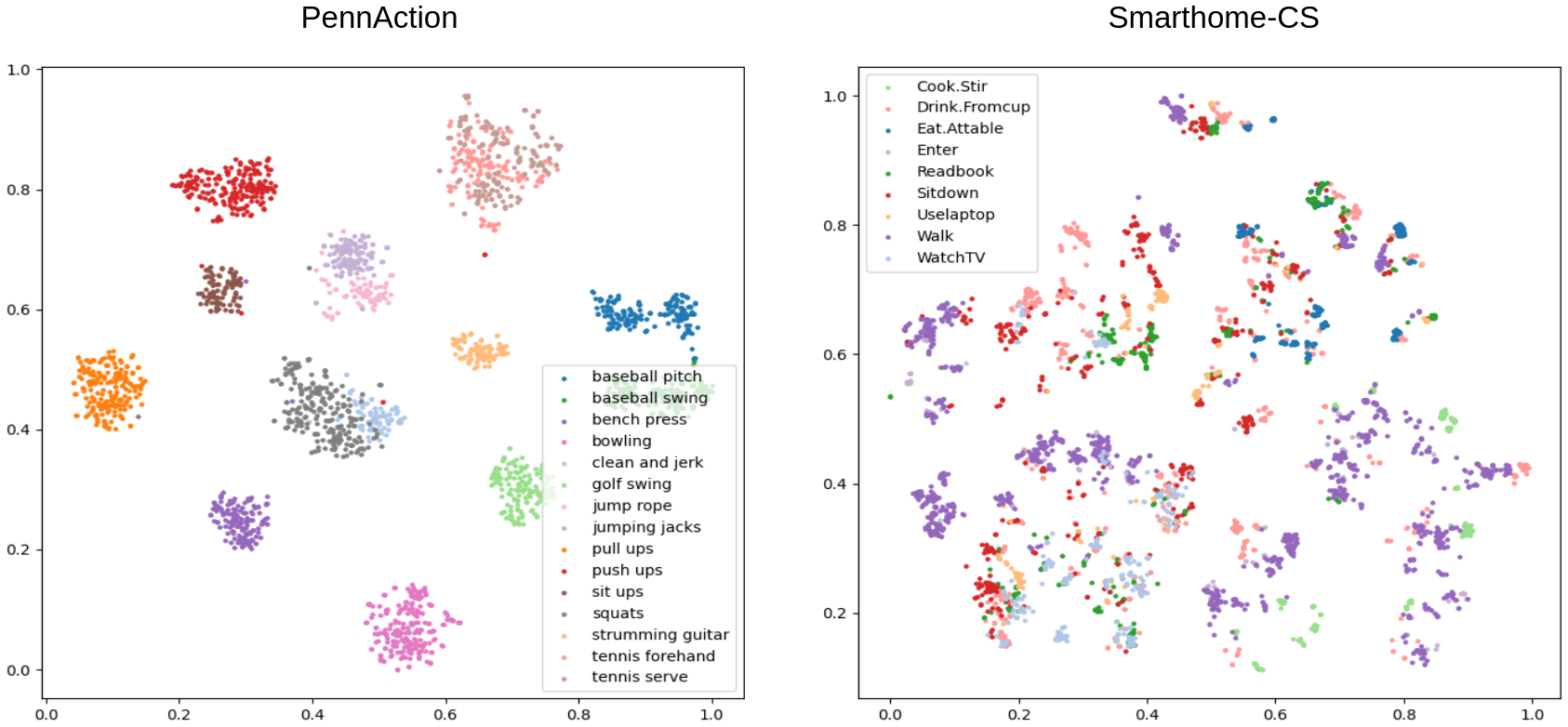}
\end{center}
   \vspace{-0.6cm}
   \caption{TSNE visualization for Penn Action and Smarthome (CS) datasets.}
\vspace{0.1cm}
\label{fig:TSNE}
\end{figure}
\begin{table*}[t]
\centering

\begin{center}
\scalebox{1}{
\setlength{\tabcolsep}{1mm}{
\begin{tabular}{  l c c c c c c }

\hline
\multirow{2}*{\textbf{Methods}} & 
\multicolumn{2}{c}{\textbf{Smarthome}} & {\textbf{Penn}} & {\textbf{UAV}}& \multicolumn{1}{c}{\textbf{NTU-10}} & {\textbf{NTU-20}} \\
& \text{CS (\%)} &\text{CV2 (\%)} & {\text{Top-1 (\%)}} & {\text{CS (\%)}} & {\text{CS (\%)}} & {\text{CS (\%)}}  \\
\hline

\text{CLIP~\cite{radford2021clip}}&\text{10.1}& \text{13.6}  & \text{63.1} & 1.6 &\text{13.8}& \text{5.1} \\
\text{X-CLIP~\cite{Ma2022XCLIP}} &\text{16.5}& \text{14.8}  & \text{72.7} & 4.8  &\text{27.6}& \text{18.8} \\
\text{ViCLIP~\cite{wang2023internvid}} &\text{14.1}& \text{14.2} & 74.3 & 1.2&\text{18.9}& \text{15.5}\\
\text{ViFi-CLIP~\cite{hanoonavificlip}} &\textbf{19.6}& \textbf{15.3}  & \text{87.1} & \textbf{5.9} &\textbf{33.4}& \textbf{23.1}  \\
\text{LanguageBind~\cite{zhu2023languagebind}} &\text{16.9}& \text{15.1}  & \textbf{90.4} & 3.7&\text{24.1}& \text{15.3}\\
\hline
\end{tabular}}}

\end{center} 
\vspace{-0.15cm}
\caption{Zero-shot transfer results (Top-1 accuracy) without re-training on 2D action classification benchmarks of Smarthome, Penn Action, UAV-Human, NTU-RGB+D 60 (only 10 classes in the test set are used) and NTU-RGB+D 120 (only 20 classes in the test set are used).}
\vspace{-0.6cm}
\label{tab_zero}
\end{table*}

\section{Experimental Analysis and Discussion}

We conduct extensive experiments to evaluate the mentioned foundation models~\cite{radford2021clip, Ma2022XCLIP, wang2023internvid, hanoonavificlip} on both action classification and segmentation tasks. We study their generalization ability by quantifying the performance improvement obtained by zero-shot learning on real-world action classification (see Sec.~\ref{sec:zeroshot}) and action segmentation (see Sec.~\ref{sec:segmentation}) datasets after visual-text pre-training. 
Subsequently, we evaluate the generalization ability of \cite{radford2021clip, wang2023internvid, hanoonavificlip, Ma2022XCLIP} by providing more analysis.

\subsection{Comparisons on Zero-shot Action Classification}\label{sec:zeroshot}

One of the advantages of visual-language model~\cite{radford2021clip} compared to classical model~\cite{arnab2021vivit, Carreira_2017_CVPR} is the application for zero-shot classification on unseen datasets. As zero-shot classification can effectively evaluate the alignment of visual and textual features, in this section, we compare the feature quality of various SoTA models~\cite{radford2021clip, Ma2022XCLIP, wang2023internvid, hanoonavificlip, zhu2023languagebind} which are trained via visual-language alignment for real-world zero-shot action classification tasks. Specifically, given a video embedding, we search its closest textual embedding (extracted using raw action labels) as the action prediction in a close-world setting. 

The results reported in Tab.~\ref{tab_zero} suggest that the original image-based CLIP~\cite{radford2021clip} model struggles with video-based tasks due to a lack of temporal consistency in the features. X-CLIP~\cite{Ma2022XCLIP} and ViCLIP~\cite{wang2023internvid} extend the CLIP model by incorporating video encoder and trained on video tasks, such as video-text retrieval~\cite{Ma2022XCLIP} and video classification~\cite{wang2023internvid} with very general video data, bring improvements but remain limited to handle fine-grained tasks (\eg, on Smarthome and UAV-Human). In contrast, ViFi-CLIP benefits from specific fine-tuning on Kinetics~\cite{k700}, which includes many actions found in the evaluation datasets, and can enhance the performance in fine-grained action classification.  However, the performance is still far away from satisfactory. The variability in viewpoints, subjects, and environmental conditions can affect the quality of visual features. 

\begin{table*}[t]
\centering

\begin{center}
\scalebox{.85}{
\setlength{\tabcolsep}{.3mm}{
\begin{tabular}{l c c c c c c c c c c }

\hline
\multirow{3}*{\textbf{Actions}} &
\multicolumn{10}{c}{\textbf{Smarthome}} \\

& \multicolumn{5}{c}{\text{CS(\%)}} & \multicolumn{5}{c}{\text{ CV(\%)}} \\

& \textit{CLIP} &\textit{XCLIP}&\textit{ViCLIP}& \textit{ViFiCLIP} &\textit{L-Bind}& \textit{CLIP} &\textit{XCLIP}&\textit{ViCLIP}& \textit{ViFiCLIP} &\textit{L-Bind} \\
\hline
\hline

\text{Eat.Attable}&\text{96.4}&\text{91.3} & \text{100.0}& 80.2 &99.2 & 100.0 & 97.3 & 100.0 & 83.7  & 100\\
\text{WatchTV}&\text{100.0} &\text{55.7} & \text{98.7} & 70.0 & 86.9& -&- & -& -  & -\\
\text{Cleandishes}&\text{6.8} &\text{68.4} & \text{51.2}& \text{50.4} &\text{48.1} & - & -& - & \text{-}  & -\\
\text{Uselaptop}&\text{0.0} &\text{44.4} & \text{53.9}& 47.2 &28.7 & 2.0 & 50.0 & 30.8 &40.4  & 40.4\\

\text{Readbook}&\text{0.0} &\text{54.2} &  \text{0.0} &\text{47.0}&\text{31.1} & 0.0 & 0.0 & 0.0 & \text{4.2}  & 0.0\\

\hline

\text{Cook.Stir}&\text{0.0} &\text{0.0} & \text{19.1}& \text{27.6} &\text{40.7} & - & -& - & \text{-}  & -\\
\text{Sitdown}&\text{0.0} &\text{5.9} & \text{0.0}& \text{16.4} &\text{5.3} & 0.0 & 0.0& 0.0 & \text{0.5}  & 2.1\\
\text{Drink.Fromcup}&\text{0.3} &\text{0.1} & \text{0.7}& \text{5.2} &\text{0.7} & 0.0 & 0.0& 0.0 & \text{0.9}  & 0.0\\

\text{Walk}&\text{0.08} &\text{1.2} & \text{58}& \text{4.7} &\text{0.5} & 0.0 & 0.0& 0.0 & \text{0.7}  & 0.0\\

\text{Enter}&\text{0.0} &\text{0.0} & \text{0.0}& \text{0.0} &\text{6.8} & 0.0 & 0.0& 0.0 & \text{0.0}  & 0.0\\

\hline
\end{tabular}}}

\end{center}
\vspace{-0.15cm}
\caption{Analysis on different actions of Smarthome using SoTA foundation models. }
\vspace{-0.25cm}
\label{tab_action}
\end{table*}

\begin{table}[t]
\centering

\begin{center}
\scalebox{1}{
\setlength{\tabcolsep}{2.mm}{
\begin{tabular}{ l c c c }
\hline
\multirow{2}*{\textbf{Methods}}& \multicolumn{2}{c}{\textbf{TSU}}&\multirow{1}*{\textbf{Charades}} \\
&{\text{ CS(\%)}} & {\text{CV(\%)}} & {\text{mAP(\%)}}\\

\hline
\hline

\text{PDAN~\cite{dai2021pdan}}  w/ CLIP~\cite{radford2021clip} &\text{16.3} &10.0 & 15.9\\

\text{PDAN~\cite{dai2021pdan}  w/ ViCLIP~\cite{wang2023internvid}} &\text{21.5} &13.4 & 16.2\\
\text{PDAN~\cite{dai2021pdan}  w/ ViFi-CLIP~\cite{hanoonavificlip}} &\textbf{28.6} &\textbf{15.9} & \textbf{16.4}\\

\hline

\rowcolor{mygray}\text{MS-TCT~\cite{dai2022mstct} w/ CLIP~\cite{radford2021clip}} &\text{5.3} & 5.7&12.7\\

\rowcolor{mygray}\text{MS-TCT~\cite{dai2022mstct} w/ ViCLIP~\cite{wang2023internvid}} &\text{15.8} & 8.2 & 16.3\\
\rowcolor{mygray}\text{MS-TCT~\cite{dai2022mstct} w/ ViFi-CLIP~\cite{hanoonavificlip}} &\textbf{21.3} & \textbf{17.3}& \textbf{16.4}\\
\hline
\rowcolor{mygray}\text{MS-TCT~\cite{dai2022mstct}} w/ I3D~\cite{Carreira_2017_CVPR} (SoTA) &\textbf{33.7} &- &\textbf{25.4} \\
\hline

\end{tabular}}}
\end{center}
\vspace{-0.15cm}
\caption{Frame-level mAP on TSU and Charades for comparison of SoTA vision foundation models with SoTA temporal modeling methods for action segmentation.}
\vspace{-0.65cm}
\label{tab_seg_sota}
\end{table}

\begin{figure}[t]
\begin{center}
\includegraphics[width=1\linewidth]{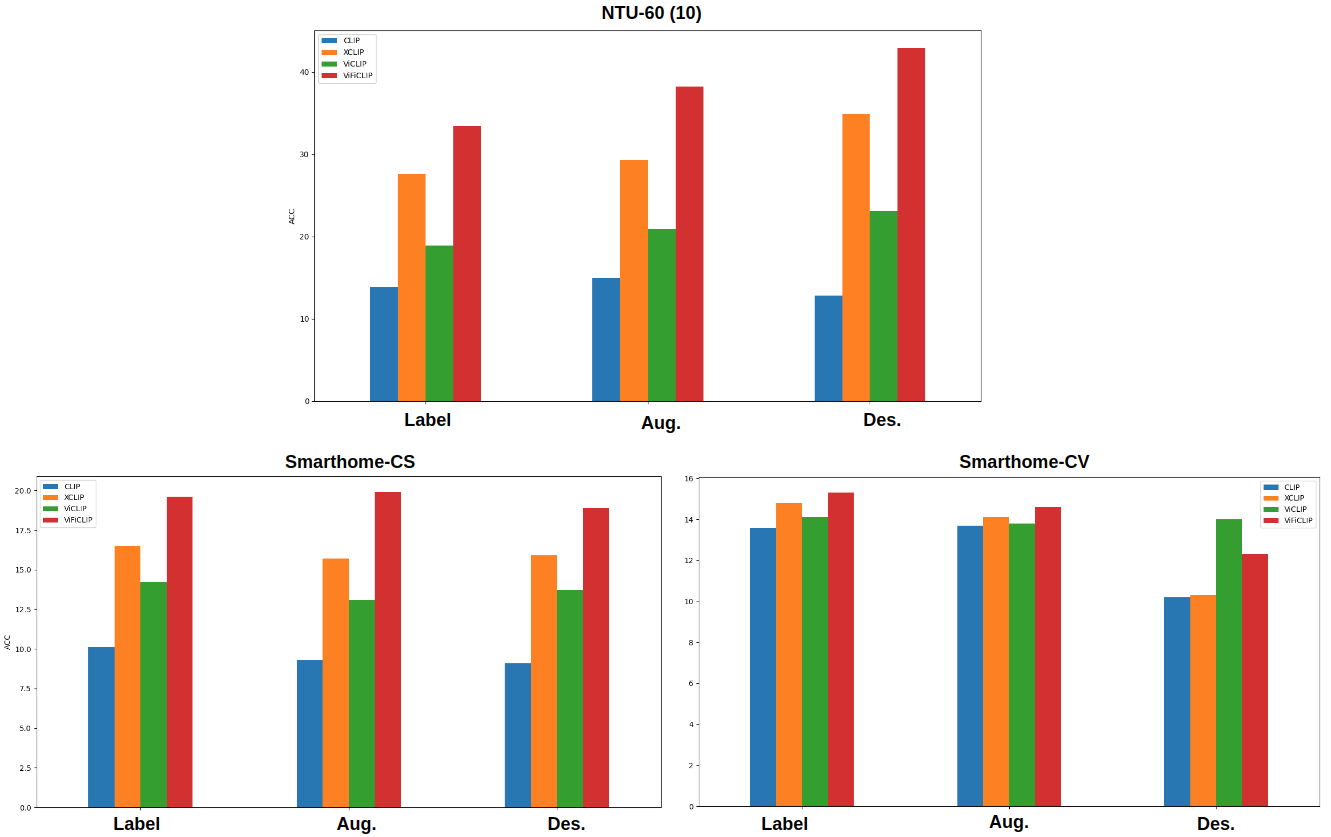}
\end{center}
   \vspace{-0.6cm}
   \caption{Comparisons of text information using raw action labels, augmented action labels (Aug.) and full action description (Des.) on NTU and Smarthome.}
\vspace{0.1cm}
\label{fig:labels}
\end{figure}
\begin{table*}[t]
\centering

\begin{center}
\scalebox{1}{
\setlength{\tabcolsep}{1.3mm}{
\begin{tabular}{  l |c c c c c c |c c c }

\hline
\multirow{3}*{\textbf{Methods}} &
\multicolumn{6}{c|}{\textbf{Smarthome}} & \multicolumn{3}{c}{\textbf{NTU-10}}\\

& \multicolumn{3}{c}{\text{CS(\%)}} & \multicolumn{3}{c|}{\text{ CV(\%)}} &\multicolumn{3}{c}{Top-1(\%)}\\

& \textit{Label} &\textit{Aug.}&\textit{Des.}& \textit{Label} &\textit{Aug.}&\textit{Des.}& \textit{Label} &\textit{Aug.}&\textit{Des.} \\
\hline
\hline
\text{CLIP~\cite{radford2021clip}}&\text{10.1}&\text{9.3} & \text{9.1}& 13.6 &13.7 &10.2 & 13.8 & 15.0 & 12.8\\
\text{X-CLIP~\cite{Ma2022XCLIP}}&\text{16.5} &\text{15.7} & \text{15.9}& 14.8 &14.1 &10.3 & 27.6&29.3 &34.9\\
\text{ViCLIP~\cite{wang2023internvid}}&\text{14.2} &\text{13.1} & \text{13.7}& 14.1 & 13.8 & 14.0&18.9 & 20.9& 23.1\\
\text{ViFi-CLIP~\cite{hanoonavificlip}}&\text{19.6} &\textbf{19.9} & \text{18.9}& \textbf{15.3} &\text{14.6} & 12.3 & 33.4& 38.2 & \textbf{42.9}\\

\hline
\end{tabular}}}

\end{center}
\vspace{-0.15cm}
\caption{Ablation study on zero-shot action classification benchmarks of Smarthome and NTU-10 with different text embeddings: original label (Label), augmented label (Aug.), and action description (Des.). }
\vspace{-0.65cm}
\label{tab_ablation}
\end{table*}

We also observe that performances are better with laboratory datasets, and are even better with the Penn-action dataset, as this is a small dataset with very few action labels.
So, these results suggest that Vision Language Foundation models are good with basic actions (similar to the web action classes), but struggle with fine-grained actions as shown in Fig.~\ref{fig:TSNE}, it is difficult to distinguish between two similar actions, just based on their labels.
It will be also interesting to perform experiments in open-world settings to verify whether performances are still good on the Penn-action dataset.

To go deeply analyze the models, in Tab.~\ref{tab_action}, we list the Smarthome classes that benefit the most and the least from the evaluated models. We find that for the actions that have very similar motions (\eg, Uselaptop vs. Readbook, Walk vs. Enter), compositional motions (\eg, Cook.Stir), and large viewpoints variations (\eg, for cross-view evaluation), the SoTA models are still limited. We can deduce from the results that more modalities (\eg, skeleton data that represents human motion) and more pre-training data are needed to further improve action recognition performance.

\begin{table*}[ht]
\centering

\begin{center}
\scalebox{1}{
\setlength{\tabcolsep}{1.8mm}{
\begin{tabular}{  l c c c c }

\hline
\multirow{2}*{\textbf{Methods}} & \multirow{2}*{\textbf{Label}}
 & \multicolumn{2}{c}{\textbf{TSU}}& {\textbf{Charades}}\\

& & {\text{ CS(\%)}} & {\text{CV(\%)}} & {\text{mAP(\%)}} \\
\hline
\hline

\text{CLIP~\cite{radford2021clip}}& 5\% & \textbf{6.2}  & 4.3 & 8.7 \\
\text{ViCLIP~\cite{wang2023internvid}}& 5\% & 3.5  &  3.3 & 10.1  \\
\text{ViFi-CLIP~\cite{hanoonavificlip}}& 5\% & 5.6 & \textbf{5.7} & \textbf{11.1} \\
\hline
\text{CLIP~\cite{radford2021clip}}& 10\% & 4.4 & 4.7 &11.1\\
\text{ViCLIP~\cite{wang2023internvid}}& 10\% & 4.0 & 3.5  & \textbf{11.6}  \\
\text{ViFi-CLIP~\cite{hanoonavificlip}}& 10\% & \textbf{6.1} & \textbf{5.8} & 11.3\\

\hline
\end{tabular}}}

\end{center}
\vspace{-0.15cm}
\caption{Transfer learning results of PDAN model on Toyota Smarthome Untrimmed (TSU) and Charades with randomly selected \textbf{5\% (top)} and \textbf{10\% (bottom)} of labeled training data after pre-training on different foundation models. }
\vspace{-0.65cm}
\label{tab_fewer2}
\end{table*}

\subsection{Comparisons on Action Segmentation}\label{sec:segmentation}

\vspace{0.2cm}

In this section, we compare the performance of the visual-language models in action segmentation tasks. As current methods for action segmentation tasks adopt a temporal model to process the continuous pre-extracted visual features on top of the untrimmed video, this experiment is to compare the representation ability of a single visual encoder of SoTA models~\cite{radford2021clip, wang2023internvid, hanoonavificlip} using their visual features with two recent temporal models~\cite{dai2021pdan, dai2022mstct} respectively.
The results in Tab.~\ref{tab_seg_sota} show that similar to zero-shot action classification, the visual representation of ViFi-CLIP is more effective than other models for segmentation tasks.
We also observe that the performances of Vision Language Foundation models are still not at the level of State-of-the-art action detection methods~\cite{dai2022mstct}.
This can be explained by the fact that these Foundation models have been trained on web videos, which are quite different from Activity of Daily Living (ADL) Videos, such as TSU or Charades.

\subsection{More Study}
In this section, we provide further analysis based on the main results.

\vspace{0.2cm}
\noindent\textbf{Can Augmenting Action Labels Improve Zero-shot Results?}
As raw action labels are too simple to fully express video content and have insufficient semantic information, we manually enrich the expression of the action labels in two levels, the augmented label and action description. Then we re-evaluate the zero-shot action classification on Smarthome and a subset of NTU-RGB+D, named NTU-10~\cite{NTU-120} with 10 selected actions. The results in Tab.~\ref{tab_ablation} and Fig.~\ref{fig:labels} suggest that the CLIP-based models are sensitive to the text embedding on NTU-10 and action description can improve the text features for zero-shot classification. However, a dataset like Smarthome, where the original labels include most information (\eg, people make coffee on the table), does not benefit from the augmentation of action labels.

\vspace{0.2cm}

\noindent\textbf{Few-shot Learning for Action Segmentation.} Few-shot transfer-learning is commendable and enables obtaining good accuracy with limited labeled data. This highlights the model practicality in real-world applications where data scarcity is prevalent. The few-shot transfer ability of our evaluated CLIP-based models on top of temporal modeling~\cite{dai2022mstct} is shown in Tab.~\ref{tab_fewer2}. The results are consistent with previous evaluation, ViFi-CLIP~\cite{hanoonavificlip} has mostly the best visual representation ability.

\begin{table}[t]
\centering

\begin{center}
\scalebox{1}{
\setlength{\tabcolsep}{2.mm}{
\begin{tabular}{ l c c c }
\hline
\multirow{2}*{\textbf{Methods}}& \multicolumn{2}{c}{\textbf{TSU}}&\multirow{1}*{\textbf{Charades}} \\
&{\text{ CS(\%)}} & {\text{CV(\%)}} & {\text{mAP(\%)}}\\

\hline
\hline

TimeChat~\cite{Ren2023TimeChat}& 2.5 & 3.4 & 14.7\\
UniVTG~\cite{lin2023univtg}& 2.4 & 3.2 & \textbf{17.7}\\
PDAN w/ ViFi-CLIP~\cite{hanoonavificlip}& \textbf{28.6} &\textbf{15.9} &16.4 \\
\hline

\end{tabular}}}
\end{center}
\vspace{-0.15cm}
\caption{Frame-level mAP on TSU and Charades for comparison of VQA methods with zero-shot action segmentation.}
\vspace{-0.25cm}
\label{tab_seg_vqa}
\end{table}

\begin{table}[t]
\centering

\begin{center}
\scalebox{1}{
\setlength{\tabcolsep}{2.mm}{
\begin{tabular}{ l c c c c c}
\hline
\multirow{2}*{\textbf{Methods}}& \multicolumn{4}{c}{\textbf{Charades}} \\
&{\text{ R@0.3}} & {\text{R@0.5}} & {\text{R@0.7}} & {\text{mIoU}}\\

\hline
\hline

TimeChat~\cite{Ren2023TimeChat}& 42.4 & 23.1 & 9.4 & 28.0\\
UniVTG~\cite{lin2023univtg}& \textbf{55.8} & \textbf{29.2} & \textbf{10.6} & \textbf{31.8}\\
\hline

\end{tabular}}}
\end{center}
\vspace{-0.15cm}
\caption{R1@IOU on Charades for comparison of VQA methods with zero-shot action segmentation.}
\vspace{-0.65cm}
\label{tab_seg_vqa_iou}
\end{table}

\noindent\textbf{Can Visual-language Model Used for Zero-shot Action Segmentation?}
For zero-shot frame-wise action segmentation, one of the solutions is to apply the zero-shot action classification on each frame, which is much more complex. In this section, we propose to leverage current VQA methods~\cite{Ren2023TimeChat, lin2023univtg} to directly generate predictions of the action boundary for a given video by asking the specific questions about the actions. We compare TimeChat~\cite{Ren2023TimeChat} and UniVTG~\cite{lin2023univtg} on Charades with event-level IoU accuracy (see Tab.~\ref{tab_seg_vqa_iou}) and we find that UniVTG~\cite{lin2023univtg} is more effective. To further compare the performance of VQA model to the temporal modeling methods using CLIP-based features like PDAN~\cite{dai2021pdan} w/ ViFi-CLIP~\cite{hanoonavificlip}, we convert the action boundary to frame-level prediction and use mAP for fair comparison. The results in Tab.~\ref{tab_seg_vqa} demonstrate the UniVTG, even without the need for re-training, can achieve better accuracy than ViFi-CLIP for Charades. However, for more complex scenarios like TSU, where multiple actions can be performed in the same video and they can be overlapped, the UniVTG model is still challenging to deal with, the two stage approaches using ViFi-CLIP features.

\subsection{Discussions and Novel Direction}
From our study, we find that current SoTA visual-language foundation model still has challenge for action recognition, the semantic gap between visual features and action descriptions makes it difficult to capture fine-grained details.
To solve the problem, we suggest to use more modalities (\eg, audio~\cite{broaden_2021_ICCV} and geometry~\cite{yang2023lac}) to complement visual information, and to design more effective temporal modeling to capture long-term temporal reasoning to improve action segmentation.
Additionally, we can also take the advantage of Large Language Models to enhance the understanding of action descriptions and to improve zero-shot classification. Finally, we believe that more comprehensive datasets from real-world and video generative model~\cite{ma2024latte} can cover a broader range of actions.
\section{Conclusion}
In this study, we evaluate state-of-the-art visual-language models for fine-grained action recognition, focusing on zero-shot action classification and action segmentation. While models like ViFi-CLIP, fine-tuned on the Kinetics dataset, demonstrated the best performance, and the VQA model UniVTG, show attracted results for zero-shot action segmentation. Our results highlight current challenges in handling complex actions and long-term temporal consistency. The findings suggest that incorporating additional modalities, such as skeleton data, could enhance model accuracy and robustness. Future research should explore integrating multi-modal data and fine-tuning strategies to improve action recognition performance.

%
%
\bibliographystyle{splncs04}
\bibliography{main}
\end{document}